\documentclass{sigplanconf}

\usepackage{amsmath}
\usepackage{graphicx}

\begin{document}

\setlength{\pdfpageheight}{\paperheight}
\setlength{\pdfpagewidth}{\paperwidth}

\conferenceinfo{CONF 'yy}{Month d--d, 20yy, City, ST, Country}
\copyrightyear{20yy}
\copyrightdata{978-1-nnnn-nnnn-n/yy/mm}
\doi{nnnnnnn.nnnnnnn}
\titlebanner{banner above paper title}        
\preprintfooter{short description of paper}   

\title{Sentiment Analysis for YouTube Comments in Roman Urdu}

\authorinfo{Tooba Tehreem}
           {National University of Computer and Emerging
           Sciences Islamabad, Pakistan}
           {Email: i191240@nu.edu.pk}
\authorinfo{Hira Tahir}
           {National University of Computer and Emerging
           Sciences Islamabad, Pakistan}
           {Email: i192071@nu.edu.pk}
\maketitle
\begin{abstract}
Sentiment analysis is a vast area in the Machine learning domain. A lot of work is done on datasets and their analysis of the English Language. In Pakistan, a huge amount of data is in roman Urdu language, it is scattered all over the social sites including Twitter, YouTube, Facebook and similar applications. In this study the focus domain of dataset gathering is YouTube comments. The Dataset contains the comments of people over different Pakistani dramas and TV shows. The Dataset contains multi-class classification that is grouped The comments into positive, negative and neutral sentiment. In this Study comparative analysis is done for five supervised learning Algorithms including linear regression, SVM, KNN, Multi layer Perceptron and Naive-Bayes classifier. Accuracy, recall, precision and F-measure are used for measuring performance. Results show that accuracy of SVM is 64 percent, which is better than the rest of the list.
\end{abstract}

\section{Problem statement}
As social networks are growing in popularity due to their quick and easy access and their low cost.  To analyze the sentiments through User
roman comments on YouTube by using roman Urdu.
\begin{figure}[htbp]
    \centering
    \includegraphics[width=0.5\textwidth]{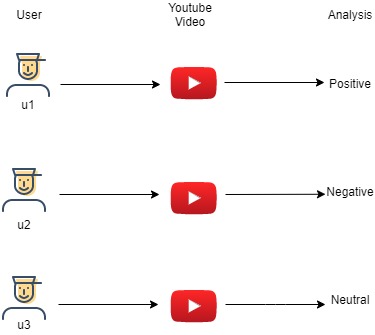}
    \caption{}
\end{figure}

\section{Introduction}
\begin{itemize}
\item	To get sentiment analysis results from comments / texts from Roman Urdu. The authors [8] did their best to
include all research done so far in the classification of the Roman Urdu. During this study, they
observe that the classification was done in two Or three grades, if it was (Negative, Positive or
Neutral). For this purpose, most researchers have used logistic regression, support vector machine and naive Bayes.
\item	Motivation: 
Since almost all Pakistanis use Roman Urdu as the language of communication, there is a lot of data on the 
Internet in Roman Urdu that can be used to analyze feelings.
\item	Background: YouTube is an area where users can view, report, rate, comment, share, upload, and bookmark on videos and 
Subscribe to the channels. The content available on YouTube contains Excerpts from television shows, movie trailers, educational, audio recordings,  videos,
Music videos, video blogs, films, and Documentaries.
The reader needs less information to sentiment. It is a study to inspect feelings, opinions, attitudes, ratings and response of users' who broadcast on social networks. A huge number of users'
comments show the current situation of feedback. It is a difficult challenge for a human to get the latest trends and summarize users' choice as there is a huge number of data on social media brings about that
needs analysis.
\end{itemize}

\section{Related work}
Much work has been done on the sentiment analysis with YouTube. Benkhelifa, et al. \cite{benkhelifa2018opinion} discussed the statement
extraction and classification of comments on cooking recipes from Youtube in real time. A real-time system has been proposed that automatically extracts and YouTube are categorized cooking recipes. After collecting the Data,
filter the comments and classify the comments based on 
SVM classifier.\\
This \cite{kaur2019cooking} focuses on the sentiment analysis of Hinglish's comments on cooking channels. Unsupervised learning, the DBSCAN technique was used to find different patterns in the comment data.\\
Yu et al. \cite{yu2013good} Suggested a method for predicting reviews from online prescription users. About
the ingredients of the recipe, recipe instructions and reviews are taken into account. The
Multi-class SVM was used to study the reliability of this information. In their study, they found
that the journal information provided the most reliable predictions.\\
Timoney et al. \cite{timoney2018nostalgic} Have been doing sentiment analysis on YouTube Videos of the best British songs since 1960.\\
Khan et al. \cite{khan2017semi} Conducted a sentiment analysis of UK film data and Amazon product review data
the semi-supervised approach. The Lexical methodology has been combined with machine learning to achieve improvements in their study of
Sentiment analysis.\\
Uysal \cite{uysal2018feature} used the various function selection methods with supervised classification techniques on YouTube
Remarks.\\
\cite{asif2019dataset} Described the decision tree sentiment analysis.\\
\cite{qutab2020sentiment}described the roman urdu text concepts.\\
This paper \cite{qamarrelationship} used to describe the emotions and behavior of human beings on social media.\\
Paper \cite{awan2021top} described the Cyprus classification and extraction.\\
 The author of \cite{nacem2020subspace} shows the improvement in automatic speech recognition.\\
 \cite{javed2020collaborative} described the health care analysis.
These papers\cite{zahid2020roman} \cite{majeed2020emotion} described mobile reviews using F1-score applying on Roman Urdu.\\
These \cite{asad2020deepdetect} \cite{javed2020alphalogger} \cite{naeem2020deep} used for detection purposes using deep learning.\\
\cite{zafar2020search} \cite{zafar2019using} \cite{zafar2019constructive}  described the game sensation.
The paper\cite{arshad2019corpus} studied about the emotion of corpus using Roman Urdu.\\
 These papers \cite{farooq2019melta} \cite{farooq2019bigdata} described energy consumption for Android studio.\\
 \cite{sahar2019towards} \cite{javed2019fairness}  described the improvement in energy consumption.
\section{Proposed approach}
The model proposed is split into four major steps. First one is, the Reviews written in Roman Urdu are gathered in the file and labelled Manually into 3 categories, i.e. positive, negative and neutral. Then Data is preprocessed. After that, features are selected. Data is divided into training and testing. Different models are used by applying different classifiers, and then the outcomes are inspected and relate. The methodology includes the steps shown in the following figure:
\begin{figure}[ht]
 \centering
  \includegraphics[width=1in]{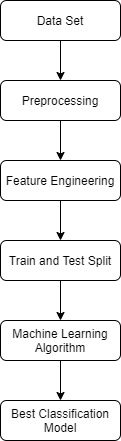}
  \caption{ Methodology Flow Chart}
\end{figure}
\subsection{Dataset}
Dataset for this study was downloaded from Kaggle. This is a labeled dataset which contains 2 columns. The 1st Column consists of Comments gathered from YouTube in roman Urdu whereas 2nd Column consists of class labels. This dataset comprises of ternary Classifications i.e. all comments are divided into different classes i.e. positive, negative and neutral. The First column is for comments, second column is of sentiment, whereas the Third column is titled as nan, which Contain null values so we removed that column from the dataset.\\
The Dataset contains 14131 unique records. Percentage of neutral comments is 48 percent, positive comments is 29 percent and negative comments is 24 percent.\\
Below bar graph in fig shows the dataset division with respect to labels.
\begin{figure}[ht]
 \centering
  \includegraphics[width=3.5 in]{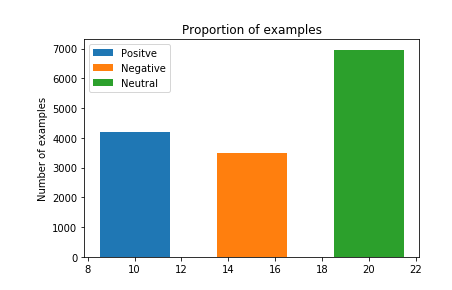}
  \caption{Proportion of Labels}
\end{figure}
\\\\\\\\\\\\\\\\
A Glimpse of the dataset is shown in the following fig having two Columns i.e. Comment and sentiment.
\begin{figure}[ht]
 \centering
  \includegraphics[width=3.5 in]{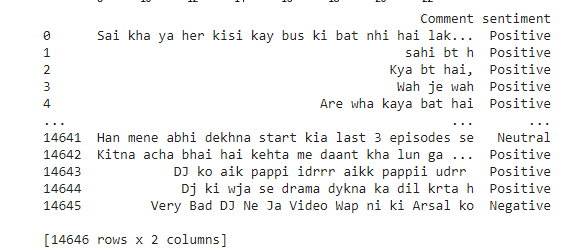}
  \caption{Glimpse of dataset before preprocessing}
\end{figure}

\subsection{Preprocessing}
Preprocessing is done in 5 steps as shown in fig:
\begin{figure}[ht]
 \centering
  \includegraphics[width=3.5 in]{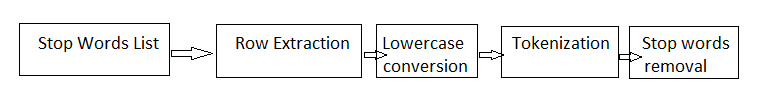}
  \caption{Glimpse of dataset before preprocessing}
\end{figure}
\subsubsection{Stop words List creation}
A list of roman Urdu stop words are created manually, which contains around 100 words.
\subsubsection{Row Extraction}
Each row is extracted containing comment from the data frame one by one in order to apply preprocessing steps on it.
\subsubsection{Lower case conversion}
Upper case letters are converted into lower case using lower () method. 
\subsubsection{Tokenization}
Each row is split into tokens of words using split () method.
\subsubsection{Removal of Stop Words}
In this step, stop words are removed from the row and remaining comment is added into a new column of a dataset with name ‘text final’ which is appended in the original dataset for further use.\\\\
It can be clearly seen in following Fig that stop words has been removed and preprocessed text is stored in the third column of the dataset. 
\begin{figure}[ht]
 \centering
  \includegraphics[width=3.2 in]{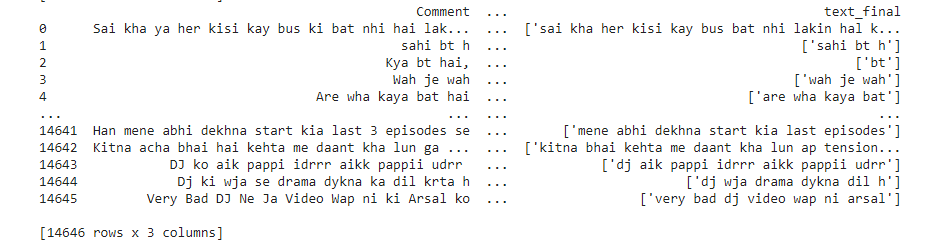}
  \caption{Glimpse of Dataset after preprocessing}
\end{figure}
\subsection{Feature Engineering}
The Purpose of this step is to convert textual data into numerical values in order to feed them to classifiers.
There are many ways to do so, but the method used in this study is ‘TF-IDF’ which stands for Term Frequency- Inverse Document Frequency.\\
In this method, frequency of each word in the data frame is calculated. A unique number and frequency is also assigned to each word. This is also known as word vectorization.
TfidfVectorizercan be easily imported from sklearn library of feature extraction.\\

from sklearn.feature extraction.text import TfidfVectorizer\\

The TF-IDF model is fitted on the whole dataset by using the method of TfidfVectorizer.Fit() which takes the preprocessed column of dataset as an input and provide a vocabulary of words with unique integer numbers attached to them as an output.\\ 
In this way, the dataset is ready to become as an input of classifiers.\\
Glimpse of frequency of each word is shown in following figure:
\begin{figure}[ht]
 \centering
  \includegraphics[width=3 in]{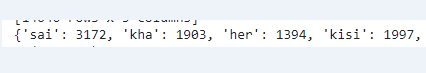}
  \caption{Word Frequency}
\end{figure}
\subsection{Train and Test Split}
The Dataset divides into two halves, one use for Training while the other use for Testing. Classification models fit on the training dataset while  the test dataset uses for making predictions. Sklearn library use for splitting the data on two halves and training-testing ratio mentione as an input for the function.80 percent of dataset provides for training, whereas predictions performed on 20 percent of the remaining dataset. Parameter for the training set to 20 percent.
\subsection{Machine Learning Algorithms}
In this study, a total of 5 ML algorithms are used which are discussed above.\\
All models are imported and run using the Sklearn python library. Selecting an appropriate model for dataset is crucial because accuracy in prediction depends very much on the model.\\
As discussed in the introductory session, below classifiers are used for this comparative study.\\
\begin{itemize}
\item	Naïve Bayes Classifier
\item	Linear Regression
\item	Support Vector Machine
\item   KNN
\item   Multi-Layer Perceptron
\end{itemize}

\section{Evaluation and Experiments}
For evaluation purpose of sentiment analysis confusion metrics is used as an aid which shows the values of true positive, true negative, false positive and false negative parameters.\\
These parameters are further used to calculate accuracy score, precision score, recall score and f1-measure score. Precision and recall combine to generate f1-score.\\
In our experiment, we imported the evaluation metrics from the sci-kit library and used them in python code. Accuracy of all algorithms is discussed individually later on in this paper.\\
All classifiers are run 10 times and results are stored in the list. Mean values are calculated for accurate and closest results.
\subsection{Naïve-Bayes algorithm}
Naive Bayesian classifiers are a family of simple "probabilistic classifiers" based on the application of Bayes' theorem with strong assumptions about independence between features. They are some of the simplest Bayesian network models, but when combined with the estimation of the core density, they can achieve a higher level of precision.\\
Confusion metrics for Naive-Bayes.
\begin{center}
\begin{tabular}{ c c c }
 127 & 141 & 53 \\ 
 33 & 594 & 72 \\  
 29 & 187 & 229    
\end{tabular}
\end{center}
\begin{figure}[ht]
 \centering
  \includegraphics[width=3 in]{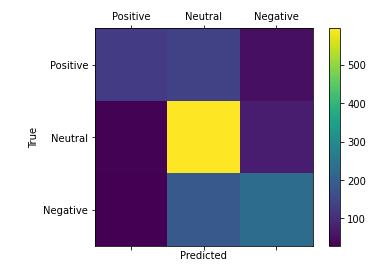}
  \caption{Naive-Bayes}
\end{figure}
\subsection{Linear regression}
In statistics, linear regression is a linear approach to modeling the relationship between a scalar response and one or more explanatory variables. The case of an explanatory variable is called simple linear regression. If there are several, the process is called multiple linear regression.\\
Confusion Matrix of Linear regression.
\begin{center}
\begin{tabular}{ c c c }
 140 & 134 & 47 \\ 
 44 & 582 & 73 \\  
 34 & 185 & 226   
\end{tabular}
\end{center}
\begin{figure}[ht]
 \centering
  \includegraphics[width=3 in]{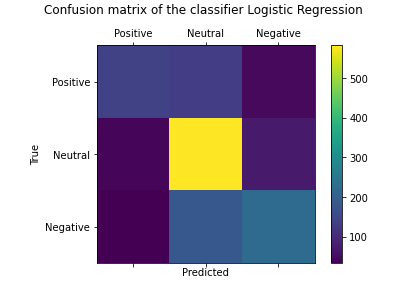}
  \caption{Linear Regression}
\end{figure}
\subsection{Support Vector Machine}
In machine learning, supporting vector machines are kept an eye on learning models using associated learning algorithms that examine data for regression analysis and classification .\\
Confusion Matrix of SVM.
\begin{center}
\begin{tabular}{ c c c }
 165 & 111 & 45 \\ 
 55 & 560 & 84 \\  
 51 & 161 & 233   
\end{tabular}
\end{center}
\begin{figure}[ht]
 \centering
  \includegraphics[width=3 in]{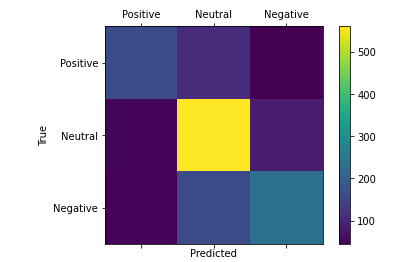}
  \caption{SVM}
\end{figure}
\subsection{KNN}
K's Nearest Neighbor (KNN) is a manageable, user friendly, flexible algorithm and one of the best in machine learning. KNN is used in a variety of applications such as image recognition, finance,political science, healthcare,video recognition  and handwriting recognition.\\
Confusion Matrix of KNN.
\begin{center}
\begin{tabular}{ c c c }
 165 & 111 & 45 \\ 
 55 & 560 & 84 \\  
 51 & 161 & 233   
\end{tabular}
\end{center}
\begin{figure}[ht]
 \centering
  \includegraphics[width=3 in]{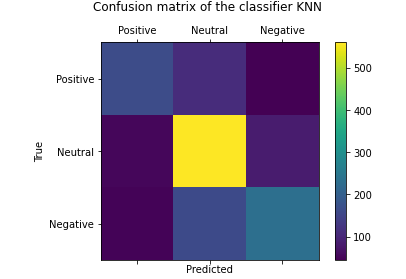}
  \caption{KNN}
\end{figure}
\subsection{MLP}
A multilayer perceptron is a kind of artificial neural network with direct feedback. The term MLP is used vague, at times loosely for any forward-looking ANN, seldom strictly to denote networks made up of multiple layers of perceptrons.\\
Confusion Matrix of MLP.
\begin{center}
\begin{tabular}{ c c c }
 156 & 98 & 67 \\ 
 129 & 433 & 137 \\  
 75 & 126 & 244   
\end{tabular}
\end{center}
\begin{figure}[ht]
 \centering
  \includegraphics[width=3 in]{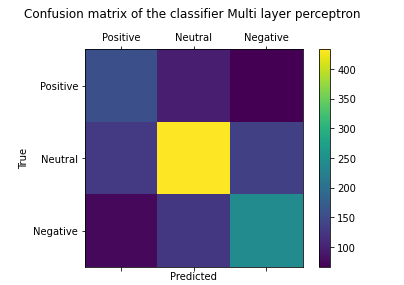}
  \caption{MLP}
\end{figure}
\section{Results:}
The Ten-fold cross validation technique is used to evaluate the performance of classifiers. The Dataset is irregularly separated into  training and testing partitions. 90 percent data are used for training and 10 percent data are used for testing. \\
\begin{figure}[ht]
 \centering
  \includegraphics[width=3.5 in]{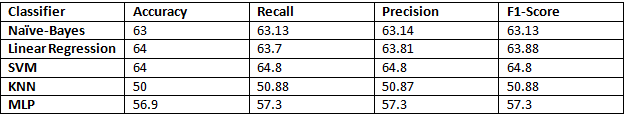}
  \caption{}
\end{figure}
\subsection*{
Accuracy graph
}
Following is the accuracy graph in figure 14.
\begin{figure}[ht]
 \centering
  \includegraphics[width=3.5 in]{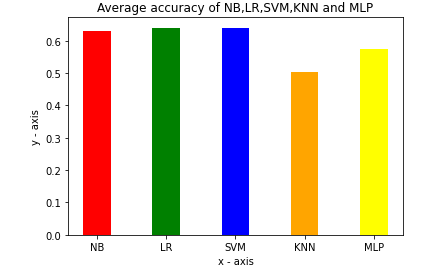}
  \caption{Accuracy Graph}
\end{figure}\\\
\subsection*{
Precision graph
} Following is the precision graph in figure 15.
\begin{figure}[ht]
 \centering
  \includegraphics[width=3.5 in]{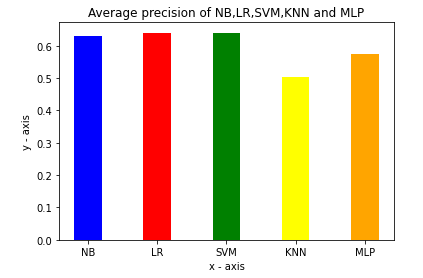}
  \caption{Precision Graph}
\end{figure}
\\\
\subsection*{
Recall graph
}
Following is the recall graph in figure 16.
\begin{figure}[ht]
 \centering
  \includegraphics[width=3.5 in]{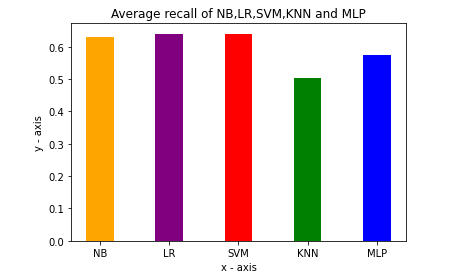}
  \caption{Recall Graph}
\end{figure}
\\\
\subsection*{
F1-score graph 
}
Following is the F1-score graph in figure 17.
\begin{figure}[ht]
 \centering
  \includegraphics[width=3.5 in]{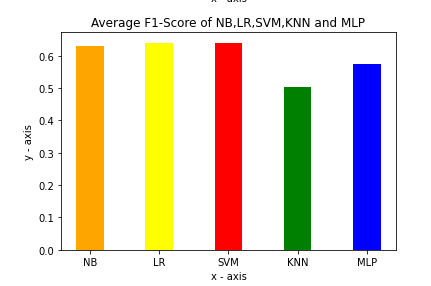}
  \caption{F1-score Graph}
\end{figure}
\\
\section{Conclusion}
In this study 5 supervised learning algorithms are applied to a dataset to predict which classifier performs better on roman-Urdu dataset. Feature extraction technique used for this study is anagram which is also known as a bag of words that contains all words in the form of tokens. Maximum features selected for this study from the feature vector are 3000. Our results show that SVM is better in terms of accuracy as relate to other four algorithms.  In future work, we will apply multiple feature extraction ability in order to upgrade the all-around precision of the models.
\bibliographystyle{plain}
\bibliography{bibliography.bib}
\end{document}